\documentclass[dvipsnames,format=sigconf]{acmart}

\AtBeginDocument{%
  \providecommand\BibTeX{{%
    \normalfont B\kern-0.5em{\scshape i\kern-0.25em b}\kern-0.8em\TeX}}}

\usepackage{booktabs} %

\usepackage{tikz}
\usetikzlibrary{arrows,calc}
\usepackage[table]{colortbl}%

\usepackage{subcaption}
\usepackage{qtree}
\usepackage{xspace}
\usepackage{multirow}
\usepackage{listings}   
\usepackage{enumitem}
\usepackage{float}
\usepackage{setspace}

\copyrightyear{2024}
\acmYear{2024}
\setcopyright{rightsretained}
\acmConference[GECCO '24 Companion]{Genetic and Evolutionary Computation Conference}{July 14--18, 2024}{Melbourne, VIC, Australia} \acmBooktitle{Genetic and Evolutionary Computation Conference (GECCO '24 Companion), July 14--18, 2024, Melbourne, VIC, Australia} \acmDOI{10.1145/3638530.3654277}
\acmISBN{979-8-4007-0495-6/24/07}

\begin{document}
\title{Guiding Genetic Programming with Graph Neural Networks}

\author{Piotr Wyrwiński}
\orcid{0000-0001-9796-5025}
\affiliation{\institution{Poznan University of Technology}
 \city{Poznan}
 \country{Poland}
}
\authornote{Corresponding author: piotr.wyrwinski@cs.put.poznan.pl}

\author{Krzysztof Krawiec}
\orcid{0000-0001-5439-3231}
\affiliation{%
 \institution{Poznan University of Technology}
 \city{Poznan}
 \country{Poland}
}

\renewcommand{\shortauthors}{}

\newcommand{\mnamens}{NUDGE} %
\newcommand{\mname}{\mnamens\xspace}

\begin{abstract}
In evolutionary computation, it is commonly assumed that a search algorithm acquires knowledge about a problem instance by sampling solutions from the search space and evaluating them with a fitness function. This is necessarily inefficient because fitness reveals very little about solutions -- yet they contain more information that can be potentially exploited. To address this observation in genetic programming, we propose Evo\mname, which uses a graph neural network to elicit additional knowledge from symbolic regression problems. The network is queried on the problem before an evolutionary run to produce a library of subprograms, which is subsequently used to seed the initial population and bias the actions of search operators. In an extensive experiment on a large number of problem instances, Evo\mname is shown to significantly outperform multiple baselines, including the conventional tree-based genetic programming and the purely neural variant of the method. 
\end{abstract}





\maketitle
\section{Introduction}

The blueprint of evolutionary algorithms assumes that the fitness function is the only means by which the search method is informed about the characteristics of a given problem instance. This design choice is inspired by natural evolution, where a species cannot improve its adaptations otherwise than by spawning randomly diversified offspring, some of which have the chance of being fitter than others. However, there is no reason to keep imposing this information bottleneck if other sources of informative guidance are available, which is relatively common in the practice of metaheuristic search algorithms. For instance, if a problem instance features constraints, one may seed the initial population with candidate solutions that comply with them; if the distributions of some variables happen to be known in advance, one may design search operators that take those distributions into account. In this study, we aim at eliciting problem-specific knowledge also from the candidate solutions themselves and from how they are being evaluated.  

As per the No Free Lunch Theorem \cite{wolpert:1997:nflto}, an optimization algorithm informed about the characteristics of a problem instance cannot perform worse on average than an uninformed algorithm. However, gathering useful knowledge about a problem and turning it into information that is `actionable' for the search policy is difficult in domains where the fitness function depends on solutions in a complex way. %
One domain with this characteristic is genetic programming (GP), where solutions are programs or other symbolic expressions that reveal their characteristics only once executed. %
Turning the effects of program execution into search guidance is difficult, but can be realized as a learnable mapping. 
To this aim, we hybridize the GP heuristics with a bespoke graph neural network (GNN) designed to generate graphs of programs. Given an instance of a GP problem represented as a set of input-output examples, the GNN is queried on it to produce a sample of GP subprograms, which is then used to seed the GP population and bias the search operators. We apply this approach to symbolic regression (SR), but it can be easily generalized to other domains. 

The main contributions of this study are (i) Evo\mname, a neuro-evolutionary method for solving SR problems (Sec.\ \ref{sec:evomethod}) and its experimental assessment on a range of SR benchmarks (Sec.\ \ref{sec:experiments}). The remaining sections comprise problem formulation (Sec.\ \ref{sec:formulation}) and the review of related works (Sec.\ \ref{sec:related}).

\section{Problem formulation}\label{sec:formulation} %

The class of problems considered in this paper is symbolic regression (SR), where the task is to construct a mathematical expression that maps a number of independent variables $x_i$ to a dependent variable $y$ so that the regression \emph{model} obtained in this way minimizes an approximation error (typically MSE) on a set of training examples $T=\{(\mathbf{x}^{(j)},y^{(j)})\}$, and prospectively generalizes well beyond this sample. SR is a special case of program synthesis from examples, where the search space is defined by the \emph{domain-specific language} (DSL) comprising the set of instructions $O$ (mathematical operators) and the set of terminals $V$, i.e. input variables and constants. Any finite tree formed by composing the elements of $O$ and $V$ is a valid program. We assume that execution has no side effects.

\section{The proposed method}\label{sec:evomethod}

\subsection{Motivations} 
\mname, the neural component of Evo\mname, is designed to address selected limitations of GP as an SR method and so complement it via hybridization. 

When framing SR (or any program synthesis) as a search or optimization problem, it is common to assume the search space to span \emph{complete} solutions (SR models), and the search algorithm to traverse that space, guided by an objective function. This formulation, typical also for GP, has nevertheless several downsides. Firstly, it largely ignores the \emph{compositional} nature of programs, i.e. that they are built of parts (instructions, subprograms) that can be combined according to the prescribed syntax rules and then reasoned about.  %
In many cases, including SR, such subprograms can be even independently executed to reveal their characteristics. Yet, from the viewpoint of a GP algorithm, a candidate solution is an opaque entity that cannot be inspected for clues about its potential for improvement. %

Secondly, working with complete solutions makes it hard to  \emph{acquire knowledge} about (i) the given instance of the problem and (ii) the domain as a whole -- by which we mean here primarily the characteristics of the underlying DSL and the considered class of problems. 
We posit that these characteristics make it hard for GP algorithms to benefit from the \emph{incremental} %
nature of the search process. Like most evolutionary metaheuristics, GP is an iterative algorithm, where the population is expected to gradually accumulate candidate solutions of increasing quality while sustaining further exploration. In this sense, the population materializes the cumulative knowledge of the search process and forms its memory. But the mechanisms offered by the evolutionary blueprint to maintain, update, and exploit that knowledge are very limited and are applicable only to complete solutions -- in most cases, those include just selecting and deselecting candidate solutions for/from the population. An evolutionary process cannot, for that instance, maintain a solution (or a part thereof, i.e. a piece of code) that is only \emph{prospectively} useful (or `interesting' in some other sense) unless it happens to perform on par with other competing solutions currently in the population. In more general terms, GP algorithms cannot \emph{reason} about the problem they are solving, even in a very rudimentary sense of this word.

To address the above limitation, multiple techniques have been proposed, like niching to protect original solutions from extinction, novelty search to promote fitness-agnostic exploration, or model-based search techniques to materialize the acquired knowledge as a probability distribution or other structure (see Sec.\ \ref{sec:related} for review). Nevertheless, hardly any of those methods attempt to explicitly elicit knowledge from the problem being solved, and very few of them looked into the compositional character of candidate solutions as a potential source of search guidance. 

\subsection{\mname}\label{sec:gnn}

The \emph{Neuro gUideD Graph sEarch} we propose in this study addresses the above limitations of GP by conducting a preliminary search on the level of solution components, rather than complete solutions. Its search state is the graph spanning all partial solutions generated so far, built gradually bottom-up using elementary components from $V$ (input variables and constants) and $O$ (operations/instructions, Sec.\ \ref{sec:formulation}). The graph provides the search algorithm with an integrated, coherent view of the problem-specific knowledge collected so far. The search is guided by a graph neural network (GNN) that has access to the syntactic and semantic information stored in the graph, as detailed in the following. %

\emph{Graph search algorithm}. 
The proceeding of \mname can be likened to a prioritized construction of the transitive closure of $V \cup O$, i.e. the graph of expressions that can be built bottom-up from the variables, constants, and operations available in the SR formulation and the provided DSL. Starting from the initial edgeless digraph $G_0=(V\cup O,\emptyset)$, the search algorithm builds in each iteration $G_{t+1}$ by expanding $G_t$ with a `layer' of new nodes and edges, where the choice of the nodes to be added is controlled by the GNN. This can be thus seen as an incremental exploration of the above-mentioned transitive closure. 

\begin{figure}[t]
    \centering
    \includegraphics[width=\linewidth]{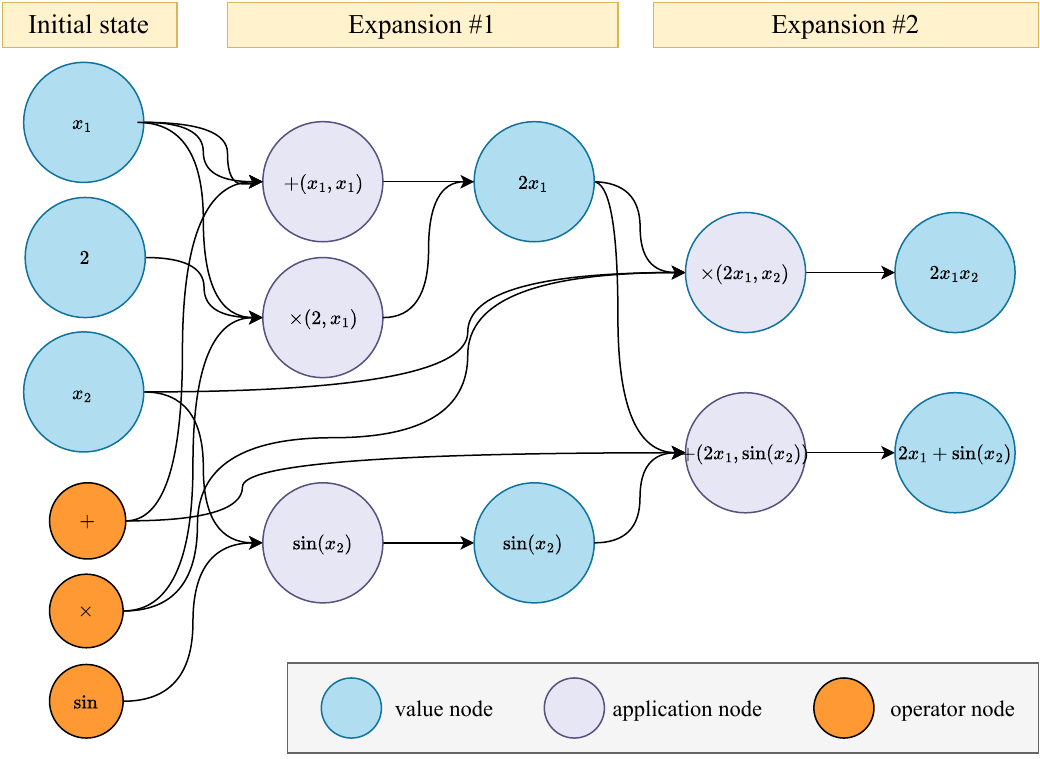}%
    \caption{An example of a graph constructed by \mname.}
    \label{fig:graph-example}
\end{figure}

More precisely, the subsequent graph $G_{t+1}$, $t>0$ is formed by expanding the previous graph $G_t=(N,E)$ with (i) \emph{application nodes} $N_a$, each representing the application of an operation from $O$ to arguments from $N$, and (ii) \emph{value nodes}, which represent the outcomes of those applications. For instance, applying the $\div$ operation to the constant nodes $8$ and $4$ already present in $N$ results in the corresponding application node labeled $\div(8,4)$ and a value node that holds the value $2$. In consistency with this convention, the initial constant and variables are also represented as value nodes (Fig.~\ref{fig:graph-example}).

The causal dependencies between the nodes in $N$ and the newly added ones are captured by adding directed edges to $E$: the added application node receives one incoming edge from an operation node ($\in O$), $k\geq 1$ edges incoming from variable, constant, and pre-existing value nodes ($\in (X \cup C \cup V)$), where $k$ is the arity of the operation, and one outgoing edge leading to the value node representing the outcome of the operation ($\in N_a \times V$). %

If a new application node produces a value $v$ that happens to be already in the graph, it will be connected to $v$ (cf. the application nodes $+(x_1,x_1)$ and $\times(2,x_1)$ in Fig.~\ref{fig:graph-example}). Otherwise, a new value node $v'$ is created. The equivalence of values $v$ and $v'$ is determined with symbolic execution: we trace back the edges connecting the nodes to the initial nodes of the graph, building so symbolic expressions $v'(\mathbf{x})$ and $v(\mathbf{x})$, and determine whether $v'(\mathbf{x}) - v(\mathbf{x}) \equiv 0$ by querying a symbolic executor from the SymPy library \cite{10.7717/peerj-cs.103}. The graph is thus \emph{minimal} in the above sense. 

\mname builds therefore multiple SR expression trees in parallel, representing them together as a graph that forms the working search state of the algorithm, alike the population in GP -- however in a more compact and integrated manner.

\emph{Search guidance with GNN}. 
Applying consecutive graph expansions in an unconstrained fashion is equivalent to the breadth-first search, which would quickly bloat the graph due to combinatorial explosion and so exhaust the available computational resources. To constrain the search space, we devise a bespoke GNN that,  given the current graph $G_t$, acts as an \emph{attention mechanism} that appoints only some application nodes as worth expanding. Technically, the GNN produces a \emph{saliency map} over the application nodes $N_a$ in $G_t$, i.e. $G_t \mapsto (0,1)^{|N_a|}$. %

To handle multiple examples in the set of data points $T$ that comes with the SR problem, the GNN is queried on each of them independently. We \emph{instantiate} $G_t$ with the $j$th example $(\mathbf{x}^{(j)},y^{(j)})$ from $T$ by feeding $x_i^{(j)}$s at the nodes in $V$ that represent the input variables and calculating\footnote{This is implemented using a cache that is maintained over the iterations of the algorithm, so it does not impose significant computational overhead.}  the dependent values in all value nodes in $G_t$. The GNN is then queried on so instantiated input graph and produces a saliency map $s_j$ over application nodes. The $s_j$s obtained for all examples in $T$ are averaged to form the average saliency map $s$. Finally, the $k$ application nodes with the highest saliency in $s$ are appointed for expansion, where $k$ is a parameter of the method\footnote{We experimented with other expansion policies, but this one worked best ultimately.}. 

A node of the graph instantiated for the $j$th data point is presented to the GNN as a vector comprising:
\begin{itemize}
    \item A one-hot-encoded categorical variable indicating the type of the node (variable, value, operation, application) (4 dimensions of node's representation vector).
    \item For operator nodes, one-hot encoded index of the operator in the DSL (11 dimensions for the set of 11 operators used in the experiments).
    \item For value nodes:
    \begin{itemize}
        \item The embedding of the instantiated value, using the 32-bitwise representation akin to proposed in \cite{Kamienny_d’Ascoli_Lample_Charton_2022}, presents each bit of the significand and exponent in the IEEE-754 single-precision floating-point representation of a number as a separate input to the model (32 dimensions). For the remaining types of nodes, the embedding is filled with zeroes.
        \item The signed difference between the value and target value $y^{(j)}$, embedded in the same way (32 dimensions).
    \end{itemize}
\end{itemize}
The complete node representation vector has thus $79$ dimensions.

\begin{figure*}[t]
    \begin{center}
    \includegraphics[width=\linewidth]{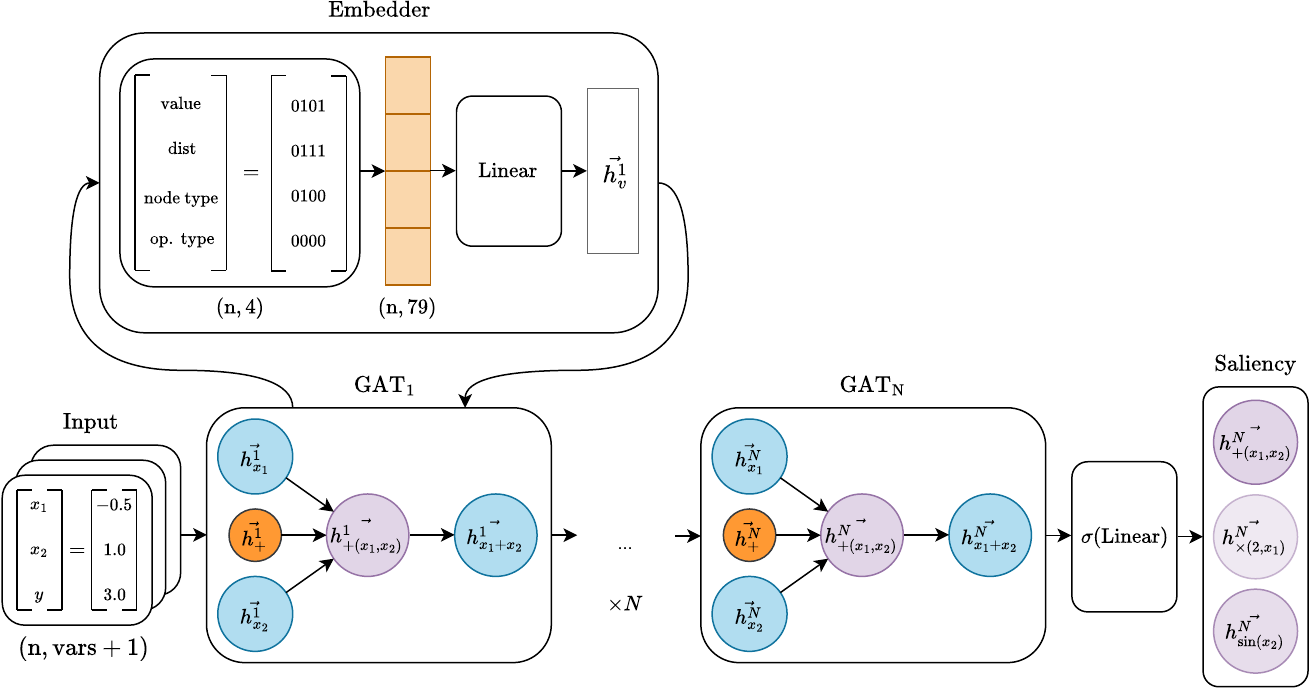}%
    \end{center}
    \caption{The architecture of the GNN used for saliency estimation in \mname; $n$: the number of examples in the dataset that specifies the SR problem instance; $N$: the number of message passing iterations and of the GAT layers of the model.}
    \label{fig:gnn}
\end{figure*}

\emph{GNN model}. 
We design a bespoke GNN architecture based on the blueprint of Graph Attention Network (GAT)~\cite{velivckovic2018graph}. GATs have proven to be effective in capturing intricate relationships within graph-structured data by assigning attention weights to neighboring nodes during several rounds of \emph{message passing} (detailed below). This attention mechanism enables the model to focus on relevant information, facilitating the extraction of intricate dependencies in the underlying graph (Fig.\ \ref{fig:gnn}). 

As most GNNs, GAT associates a working \emph{state} variable (vector) $h_v$ with each node $v$ of the input graph $G_t$, which is first initialized with the information retrieved from  $v$, then iteratively updated in consecutive rounds of message passing, and ultimately used to determine the per-node output of processing. We initialize $h_v$ by mapping the 79-dimensional node representation vector (detailed above) through a linear layer featuring 128 units.  

Then, the message passing is conducted with three GAT layers, each handling one iteration of this process and updating the state $h_v$ of every node $v$ in the graph. 
Each GAT layer is equipped with $4$ heads, applied independently to node states of dimensionality $128$ and generating a $32$-dimensional vector.
The vectors produced by heads are then concatenated to maintain the node states' dimensionality of $128$ and then passed to the next layer.
The sum aggregation operator was utilized to aggregate messages from neighboring nodes. The Exponential Linear Unit \cite{clevert2015fast} was used as the activation function in each GAT module. At each layer, nodes exchange information with their neighbors using messages, allowing the model to refine its representation of each node based on contextual information from the surrounding neighborhood. %

After message passing, the final state $h_v$ is mapped through an output layer for each graph node independently. The output layer features a single unit with the sigmoid activation function and synthesizes the information accumulated during the message-passing process, allowing for binary classification of each node in the graph, required by the training procedure detailed below. 

We train the GNN on a set of SR expression trees sampled from a given DSL (of height up to 6 in the experiments).
The training is conducted in a supervised mode, which in our particular context can be seen as \emph{imitation learning}: knowing the target program for a given SR problem instance allows us to determine the \emph{trajectory} that should be traversed by the graph expansion process to build that program. In each iteration of expansion, we identify the subexpressions (or, more precisely, application nodes) that are present in the target solution and those that are absent. This determines the desirable selection of application nodes that should be performed by the GNN, i.e. their partitioning into the positive class (nodes that should be selected) and the negative class (nodes that should not). Based on that, we define a binary target saliency vector. The actual saliency returned by the output layer of the GNN model is then confronted in training with this target saliency vector using the binary cross-entropy loss function. 
More details on this process are given in Sec.\ \ref{sec:gnn-training}. 

Because saliency values are strictly positive, \mname, thanks to the systematic nature of its search policy, is guaranteed to ultimately converge, i.e.\ construct a value node that corresponds to the sought solution (provided sufficient computational resources and assuming that the SR model can be expressed using the available operators and constants). This feature distinguishes it from some other approaches to program synthesis (including those based on large language models), which do not search the solution space systematically. The generated solutions are also \emph{syntactically correct by design} because the graph nodes follow the rules of the grammar of the underlying DSL. Arguably, this does not buy one much when considering SR problems, where all expressions have the same type and the only syntactic constraint is the arity of operators in $O$. Nevertheless, for DSLs with richer type systems, this characteristic is yet another advantage over the grammar-agnostic approaches.

\subsection{The Evo\mname hybrid}\label{sec:method}

By hybridizing \mname with tree-based GP, we expect them to complement each other. 

On one hand, \mname is better informed about the problem than the conventional GP, because it can (i) directly inspect the training examples, (ii) observe, in consecutive iterations of graph expansion, how the expressions built bottom-up from input variables `respond' to the input examples, and (iii) it has access to the error committed by those expressions with respect to the target variable $y$. But it struggles to scale well: even though only $k$ application nodes are expanded in each iteration, querying the GNN becomes more costly in consecutive iterations and the guidance it produces tends to be less accurate (i.e. it becomes more challenging for the GNN to produce well-guiding saliency for large graphs). %

On the other hand, the evolutionary search features characteristics that cannot be easily attained by \mname,  in particular the population-based, parallel search, and the possibility of `grafting' code pieces from one candidate solution to another using the crossover operator. 

Based on these observations, we adopt the following general blueprint of hybridization: given an SR problem, we first invoke \mname and allow it to perform $h$ expansions of the graph. The final graph $G_h$ produced by \mname is then used to populate a \emph{library} $L$ of subexpressions, to be later sampled from by the GP search. Technically, $L$ is obtained by traversing all the value nodes in $G_h$ and forming the corresponding expression trees. For instance, the graph shown in Fig.\ \ref{fig:graph-example} would translate into a library comprising $x_1$, $2$, $x_2$, $2x_1$, $\sin(x_2)$, $2x_1x_2$, and $2x_1+\sin(x_2)$.  

Finally, we devise the following library-based operators to be used in the GP phase of Evo\mname: 
\begin{itemize}
    \item \emph{Library-based initialization}.  
    If the intended population size $m$ is smaller than $|L|$, all expressions from $L$ are copied into the population and the missing $m-|L|$ candidate solutions are generated using the default initialization operator (the ramped half-and-half in the implementation used in Sec.\ \ref{sec:experiments}). Otherwise, we sample $m$ programs from $L$ without replacement.
    \item \emph{Library-based mutation}. This operator mimics the subtree-replacing mutation commonly used in tree-based GP. It first selects at random an expression node in the parent solution, and then replaces the subtree rooted in that node with the expression drawn at random from $L$. 
\end{itemize}
More details on these search operators are given in Sec.\ \ref{sec:experiments}.

\section{Related work}\label{sec:related}

The proposed approach relates to a number of earlier works, primarily through its most distinctive characteristics: (i) reliance on a library of solutions obtained prior to an evolutionary run and (ii) involvement of a neural network in the synthesis process. 

With respect to the \textbf{use of the library} of pre-computed subprograms, \mname can be likened to methods that maintain repositories of code pieces, often referred to as \emph{archives}. Most of such methods fill the libraries with code pieces collected throughout an evolutionary run. One of the earliest attempts of this kind was the work by Rosca and Ballard \cite{rosca:1996:aigp2} who proposed a sophisticated mechanism for assessing subroutine utility, and entropy for deciding when a new subroutine should be created. Haynes \cite{Haynes:1997:adskr} integrated a distributed search of genetic programming-based systems with `collective memory', albeit only for redundancy detection. Other approaches involving some form of a library include the reuse of assemblies of parts within the same individual \cite{hornby_alife02_s} and explicit expert-driven task decomposition using layered learning \cite{bajurnow:2004:llfegsbisp}. In a few works, the archives created in this way are reused in separate evolutionary runs applied to other problems, the notable example being \emph{run transferable libraries} \cite{ryan:2004:GPTP}.  %
\mname stands out from these approaches by relying on a library that has been precomputed before an evolutionary run. Such proceeding was relatively rare in GP-based approaches and used only occasionally, e.g. in semantic backpropagation \cite{Pawlak:2014:ieeeEC}. But even more importantly, the library generated by the GNN in \mname is conditioned on a given problem and thus tuned to it, while most of the past works relied on 'generic' libraries. 

By involving a \textbf{neural network}, \mname borders with the methods developed within \emph{neural program synthesis} that has seen a rapid acceleration in recent years thanks to advancements in deep learning. One of the earliest attempts of this kind was the DeepCoder by Balog et al. \cite{2016arXiv161101989B}, where a neural network was trained to map the input-output examples provided in the program synthesis problem to the \emph{probability distribution} of instructions to be used in the synthesized programs. DeepCoder uses that network to first query it on a given program synthesis problem, to obtain the probability estimates. Next, a search algorithm uses those estimates to parameterize its search policy, i.e. prefer some instructions to others. When combined with systematic breadth-first search and other search algorithms, DeepCoder observed manyfold speedups, ranging from 2x to 907x, depending on the assumed maximum program length and available computational budget. In a more recent work \cite{Liskowski:2018:GECCOa}, DeepCoder has been hybridized with a GP and shown to boost the efficacy of evolutionary program synthesis. 

On a conceptual level, DeepCoder is similar to \mname in being predicated on the assumption that useful information about the program to be synthesized can be obtained from the training examples \emph{in a direct way}, without resorting to an uninformed trial-and-error search. Another similarity is that it uses a neural network model to `prime' a subsequent search algorithm based on the characteristics of the problem being solved. However, it provides guidance only on the level of individual instructions (in abstraction from the context), while \mname supplies the GP run with complete subprograms. Also, the neural architecture proposed by Balog et al. \cite{2016arXiv161101989B} was very basic by today's standards: a feedforward layered network, with the input layer of appropriate size to accommodate for the information on input-output examples.  %
In contrast, \mname engages more sophisticated GNNs, which are only occasionally used for program synthesis and related tasks -- a notable exception being an interesting work on message passing for theorem proving~\cite{Paliwal_Loos_Rabe_Bansal_Szegedy_2019}, where however a GNN was used to evaluate entire graphs representing abstract syntax trees, not individual graph nodes. 

In engaging a \textbf{neural network within a GP algorithm}, our study is relatable to a range of other works like Neural Program Optimization \cite{Liskowski:2020:GECCO} and [some references anonymized for review]. %
However, to the best of our knowledge, this is the first study that uses a graph representation and a graph neural network to aid a GP algorithm.  

In a broader context, the recent advances in deep learning opened the door to purely neural generative approaches to program synthesis, and in particular to SR, in which a neural model observes the training data and directly produces the formula as a sequence of symbolic tokens. While several architectures of this kind attained impressive performance on multiple benchmarks \cite{Biggio_Bendinelli_Neitz_Lucchi_Parascandolo_2021, Kamienny_d’Ascoli_Lample_Charton_2022}, the generative approach has several limitations, which resonate with those currently pertinent to large language models (LLMs): it cannot guarantee the syntactic correctness of produced formulas, lacks transparency, and may fail to generalize beyond the training set, because, being essentially a sophisticated model of a conditional probability distribution, it tends to interpolate between the training samples, rather than extrapolate beyond them. \mname addresses those limitations, by forcing the algorithm to gradually construct a formula in accordance with the adopted grammar of expressions. This causes the resulting formulas to be syntactically correct by construction.  

Last but not least, by intertwining neural inference with symbolic processing, \mname subscribes to the class of \emph{neurosymbolic approaches}, which recently experienced a substantial revival thanks to increasing ease with which deep learning architectures can be combined with symbolic representations --- see, for instance, \cite{Garcez_Lamb_2020} and \cite{NSAI2022, DBLP:series/sbcs/ShakarianBSXP23} for reviews of state-of-the-art in this area.

\section{Experiments}\label{sec:experiments}

The overall goal of the following experimental analysis is to establish whether the guidance provided by \mname makes Evo\mname more efficient at solving symbolic regression tasks.

\subsection{The sample of SR problems}\label{sec:problems}

To provide training data for the GNN in \mname (Sec.\ \ref{sec:gnn-training}) and assess the in-sample generalization capacity of Evo\mname (Sec.\ \ref{sec:results}), we prepare a collection of SR problems by sampling expressions involving from 1 to 6 input variables, constants 0, 1, 2, 3 and $\pi$, binary operators $+, -, \times, \div$ and functions $\sqrt{x}, x^2, x^3, \sin, \cos, \log, \exp$. For each SR expression $p$ obtained in this way, we sample a $n=30$ data points $T=\{(\mathbf{x}^{(j)},y^{(j)})\}$, $j\in[1,n]$, drawing the values of independent variables $x_i$ uniformly from the interval $[1, 5]$ and setting $y^{(j)}=p(\mathbf{x}^{(j)})$. The tuple $(p,T)$ obtained in this way forms an instance of an SR problem. 

By allowing this procedure to generate expressions of height up to $6$, we obtained 1032 SR problems, which were then randomly partitioned into a \emph{training set} of 522 problems and \emph{test set} of 510 problems.

\subsection{GNN training}\label{sec:gnn-training}

The GNN was trained on the above training set using the procedure described in Sec.\ \ref{sec:gnn} until termination by a stopping condition based on the stagnation of loss function on the validation set comprising 60 SR problems set aside from the training collection. We used the Adam optimizer~\cite{DBLP:journals/corr/KingmaB14} with the learning rate $0.001$. Training required respectively 24 hours of computation on the NVIDIA DGX machine with 8 GPUs. Let us emphasize that this is a one-off process: once trained, the same single instance of the GNN was used in all experiments reported in the following.

\subsection{Configurations of compared methods}\label{sec:method-config}

The common blueprint for all configurations considered in the following is a generational GP algorithm equipped with initialization, mutation, and search operators. The parameterization of Evo\mname is almost identical to that of GP, except for the initialization and mutation operators (Sec.\ \ref{sec:method}). Other than that, all configurations use populations of size 1000 evolving for 50 generations. Each generation starts with evaluating solutions with the fitness function (mean square error, MSE), followed by selecting parent solutions with a tournament selection (tournament size 7), crossing them over with one-point crossover with probability 0.8 (otherwise copying), and subjecting both resulting children to mutation with a probability 0.2. Offspring solutions that exceed height 13 are discarded and replaced by their parents. The outcome of a run is the solution with the lowest MSE found throughout the run. 

A run of Evo\mname on a problem $(p,T)$ comprises querying \mname on the set of data points $T$ to obtain the library and then running GP with the library-based search operators (Sec. \ref{sec:method}). To generate a library, we allow \mname to perform $h$ graph expansions, with the saliency mechanism expanding $k=5$ graph nodes in each iteration, and then populate the library with trees collected from the graph. Because $h$ determines the height of subtrees available in the library, this parameter is a strong determinant of search performance. Therefore, we conduct experiments for $h=1$, $2$, and $3$, and design corresponding control configurations of GP, to assure the fairness of comparison.  

The statistics on the resulting libraries, averaged over the 510 testing problems, are shown in Table \ref{tab:lib-sizes}. Notice the relatively large ranges of library sizes, reflecting the variability of GNN's response to problem instances.  

The mutation operator used in GP uniformly draws a node in the parent tree and replaces the subtree rooted in that node with a subtree generated as follows: first, $h'$ is drawn uniformly from the $[0,h]$ interval, then a random tree of height $h'$ is generated using the `grow' method and grafted at the selected node. The library-based mutation used by Evo\mname closely mimics this operation, except for the fact that we draw a random subtree of height $h'$ \emph{from the library}. %

Concerning initialization, GP starts with the population filled with 1000 candidate solutions generated using the ramped half-and-half method \cite{koza:book}. For Evo\mname, all trees from the library $L_h$ are placed in the initial population, and the remaining $1000-|L_h|$ solutions are generated with ramped half-and-half  (cf. Sec.\ \ref{sec:method}).

\begin{table}[t]
  \caption{The statistics of library sizes, averaged over libraries produced by the GNN for all testing problems.}\label{tab:lib-sizes}
  \begin{tabular}{llccc}
    \toprule
    Metric & \hfill  & $h=1$ & $h=2$ & $h=3$ \\
    \hline
    Mean           &   & 108 & 312 & 595 \\
    Std. deviation &   &  \phantom{1}27 &  \phantom{1}67 & 160 \\
    Minimum &   &  \phantom{1}84 &  \phantom{1}84 & \phantom{1}84 \\
    Maximum &   &  213 & 489 & 891 \\
    \bottomrule
\end{tabular}
\end{table}

We compare the following configurations of Evo\mname:
\begin{itemize}
    \item $I$: uses only the library-based initialization, 
    \item $M$: uses only the library-based mutation  (probability 0.2),  
    \item $MM$: uses the library-based mutation or the baseline mutation, with a 50/50 chance (thus effectively invoking each of them with probability 0.1).   
\end{itemize}
We also test configurations that use the above operators in combination, dubbed $IM$ and $IMM$. The main baseline configuration is the GP; we also attempt solving the problems using \mname alone. %

The software implementation is based on the DEAP library \cite{DEAP_JMLR2012}.

\subsection{Results}\label{sec:results}

Table \ref{tab:success-rate} summarizes the success rates of compared methods, across the 510 test problems from the base collection. Success is defined as producing an SR model with the MSE on the testing set < $10^{-10}$. Notice that such a model can be occasionally found already in the initial population; while for GP such events are due to sheer luck, for Evo\mname variants that use the GNN-informed initialization operator ($I$) they should be attributed to the guidance learned by the GNN. The row of the table labeled \mname presents the percentages of Evo\mname runs that benefited from this property (as terminating evolution at the very beginning is equivalent to running \mname alone). %

For the fairness of comparison, we juxtapose the methods with respect to the $h$ parameter, which both in GP and Evo\mname determines the maximum height of the subtrees inserted into parent programs in mutation and used to initialize the population. Therefore, the configurations compared in columns use search operators that have very similar characteristics in terms of the expected size and shape of subtrees inserted into candidate solutions. 

\begin{table}[t]
  \caption{Success rates (percentage of successful runs out of the 510 test problems) for various heights $h$ of the subtrees inserted by the initialization and mutation operators. %
  }\label{tab:success-rate}
  \begin{tabular}{llccc}
    \toprule
    Method & \hfill  & $h=1$ & $h=2$ & $h=3$ \\
    \hline
    GP & & 28.49 & 28.09 & 31.43 \\
    \mname & & \phantom{1}1.96 & 11.20 & 13.75 \\
    \hline
                & $I$      & 30.45 & 34.38 & 43.42 \\
                & $M$      & 33.20 & 40.86 & 40.86 \\
    Evo\mnamens & $MM$     & 33.60 & 37.13 & 39.29 \\
                & $IM$     & 34.97 & 41.45 & 43.42 \\
                & $IMM$    & 33.01 & 39.69 & 40.86 \\
    \hline
                & $I$      & 30.45 & 33.99 & 37.72 \\
                & $M$      & 33.20 & 39.88 & 39.49 \\
    EvoRnd      & $MM$     & 33.60 & 33.60 & 36.54 \\
                & $IM$     & 34.97 & 40.28 & 37.92 \\
                & $IMM$    & 33.01 & 37.72 & 38.31 \\    \bottomrule
\end{tabular}
\end{table}

Evo\mname systematically outperforms GP for all considered values of $h$. The gain resulting from the use of informed search operators increases with $h$, reaching roughly 10 percent points for some configurations. Relying on the informed mutation ($M$ and $IM$) leads to a noticeably better success rate than using both the informed and uninformed (GP) mutation ($MM$ and $IMM$). Overall, $IM$ is quite clearly the best configuration across the $h$ values considered here.  

However, the observed differences between Evo\mname and the corresponding GP configurations are not only due to the problem-specific guidance provided by the GNN, but also to the fact that the overall, problem-agnostic distribution of subtrees in libraries is different than the distributions used by the initialization and mutation operators in the baseline GP. In other words, in addition to the \emph{problem-specific bias} that we intend to convey to the search operators, there is also an unknown amount of problem-independent \emph{method bias}. 

To delineate the former from the latter, we introduce additional control setups, dubbed \emph{EvoRnd}, where search operators in Evo\mname use a `wrong' library: when solving a problem $(p,T)$, we query \mname not on $T$, but on $T'$ coming from another problem $(p',T')$. This is technically realized by randomly permuting the ordering of the 510 libraries produced for testing problems so that their pairing with problems is incidental\footnote{Technically, we group the SR problems by arity (the number of input variables) and permute those groups independently, to make sure that the problem arity is equal to the number of input variables in the library.}.  

Comparison of Evo\mname and EvoRnd in Table \ref{tab:success-rate} indicates that indeed the configurations of EvoRnd are on average better than the $h$-corresponding configurations of GP, signaling that the method bias alone contributes positively to the success rate. We hypothesize that the method bias has two main constituents. Firstly, all training and test sets come from the same distribution: the problems have been generated by systematically enumerating expressions and only then randomly split into the training and test sets. This allows Evo\mname to adapt to this overall distribution, which GP is not capable of. 
Secondly, recall that the libraries disallow semantic duplicates: all expressions with the same semantics in the graph produced by the \mname (e.g. $x_1+x_1$ and $2x_1$ in Fig.\ \ref{fig:graph-example}) are collapsed to the same value node, which is then represented by a single subprogram in the library. This changes the distribution of subtrees used by search operators in a way that might be favorable for search efficiency. 

Nevertheless, EvoRnd is overall substantially worse than that of Evo\mname, which indicates that \mname manages to convey problem-specific knowledge to the GP run in a way that makes it more effective. The only exception from this is the case of $h=1$, where EvoRnd attains the same performance as \mname. The reason behind this is, however, purely technical: we first expand the graph, then check if the maximum height $h$ has been reached, and only then query the GNN for saliency. Therefore, for $h=1$, the library contains \emph{all} subtrees of height 1 and is thus the same for all problems, so randomly re-assigning libraries to problems does not change anything. These Evo\mname configurations are thus strictly speaking \emph{uninformed}, and they perform better than GP only thanks to the above-mentioned method bias. 

\begin{table}[t]
  \caption{Run times of the best performing Evo\mname IM configuration and the baseline configurations, averaged over all runs (i.e. both successful and unsuccessful ones).
  }\label{tab:times}
  \begin{tabular}{llccc}
    \toprule
    Method & \hfill  & $h=1$ & $h=2$ & $h=3$ \\
    \hline
                   & \mname & \phantom{1}0.83 &  \phantom{1}2.94 &  \phantom{1}6.35 \\
    Evo\mname $IM$ & GP search          & \phantom{1}7.98 &  \phantom{1}9.81 & 10.33 \\
                   & Total              & \phantom{1}8.81 & 12.76 & 16.68 \\
    \hline
    GP & & \multicolumn{3}{c}{\phantom{1}9.77} \\
    \bottomrule
\end{tabular}
\end{table}

The purely neural \mname method attains a much worse success rate than  Evo\mname and GP, primarily because most of the problems in the testing set require expression trees with heights greater than $3$. However, proceeding with further graph expansions does not help its performance significantly: when allowed to run much longer than any other configuration reported in Table \ref{tab:success-rate}), i.e. for 60 seconds, it achieves $14.93\%$, improving by roughly 1 percent point only on the $13.75\%$ for $h=3$. The reason is that \mname struggles to scale when the consecutive expansions become increasingly costly due to querying the GNN on larger and larger graphs. This causes its success rate to stagnate, and further expansions only occasionally lead to finding solutions. This indicates that hybridizing neural guidance with evolutionary search is beneficial from both perspectives. 

Querying the GNN and converting the resulting graph to a library incurs measurable computational overhead compared to GP. To quantify it, we measured the average times required by the \mname component and the evolutionary search for the best-performing $IM$ variant of Evo\mname and juxtapose them in Table \ref{tab:times} with the execution times of GP and \mname. As expected, greater values of $h$ lead to longer runtimes of both \mname and GP search -- the latter because higher expressions in the library lead to more tree nodes being injected into the population in initialization and mutation, which in turn causes an increase of the average tree size and higher computational cost of evaluating and manipulating such candidate solutions. Nevertheless, the total execution time of Evo\mname is still not even double the runtime of GP.

\subsection{Performance on other benchmarks}\label{sec:results-benchmarks}

To assess the out-of-sample performance of Evo\mname, we confront it with the AI Feynman suite of regression problems \cite{Udrescu_Tegmark_2020}, a collection of equations from the \emph{Feynman Lecture on Physics}. To make this suite compatible with our configuration, we removed from it three problems that used the $\arcsin$ and $\tanh$ functions (absent in our instruction set), ending up with 97 problems.

\begin{table}[t]
  \caption{Success rates on the AI Feynman suite of benchmarks for various heights $h$ of the subtrees inserted by the initialization and mutation operators. %
  }\label{tab:feynman}
  \begin{tabular}{llccc}
    \toprule
    Method & \hfill  & $h=1$ & $h=2$ & $h=3$ \\
    \hline
    GP & & 21.65 & 22.68 & 21.65 \\
    \mname & & \phantom{1}4.12 & \phantom{1}5.15 & \phantom{1}7.22 \\
    \hline
                & $I$      & 20.62 & 22.68 & 25.77 \\
                & $M$      & 23.71 & 27.84 & 24.74 \\
    Evo\mnamens & $MM$     & 23.71 & 28.87 & 30.93 \\
                & $IM$     & 24.74 & 30.93 & 26.80 \\
                & $IMM$    & 21.65 & 28.87 & 25.77 \\
    \bottomrule
\end{tabular}
\end{table}

The success rates, shown in Table \ref{tab:feynman}, are for all configurations systematically lower than those in Table \ref{tab:success-rate}, which was expected due to the independent nature of this problem suite. However, the relationships between configurations remain largely the same as in Table \ref{tab:success-rate}, except for Evo\mname $IM$ experiencing low performance for $h=3$. Nevertheless, Evo\mname maintains the upper hand compared to GP and \mname: the best configurations of the method outperform GP by almost 10 percent points, similarly as in the previous experiment. We find this result encouraging, given that a substantial fraction of problems in the AI Fenynman collection diverges in characteristics from the problems used in our training set, among others in the domains of the input variables $x_i$.

\section{Discussion and Conclusions}

We have shown that an evolutionary algorithm can be effectively and easily provided with search guidance based on the knowledge acquired from a problem instance by a neurosymbolic system based on a graph neural network. The resulting Evo\mname hybrid systematically outperforms both its constituents, showing a synergy between them. This has been demonstrated empirically on the domain of symbolic regression, for both in- and out-sample scenarios. Nevertheless, 
\mname can be potentially applied to domains beyond SR, and beyond program synthesis as such, for domains in which partial and complete candidate solutions can be represented as graphs. 

We find it particularly important that \mname, the neural component of the method, has a deep insight into the nature of the SR domain, in being able to trace the execution of symbolic expressions, examining the effects of that process on the data being processed, and relating that information to the goal of the search process, represented as the dependent variable. It is critical for our methods what part of the problem-specific knowledge obtained in this way is being passed to the evolutionary search and how, and this is the subject of our ongoing work on the method. The particular way of hybridizing \mname with GP we used here intervenes only minimally in the evolutionary pipeline by redefining the source of subtrees used by the search operators. This has the advantage of allowing for direct side-by-side comparison with the purely evolutionary configurations. On the other hand, this only scratches the surface of possible ways in which an evolutionary search can be guided with the kind of knowledge that can be gathered by \mname. For instance, when constructing the libraries, we are currently ignoring the saliency values, which can convey more nuanced information about the usefulness of particular subexpressions. Another option would be to train the GNN in combination with the GP with reinforcement learning, treating the GNNs choice of the library as an action and the resulting outcome of the GP run using that library as a reward for that action (as, arguably, in the current Evo\mname, the GNN is trained in a way that is agnostic about the specific `needs' of the evolutionary algorithm).

\begin{acks}
  We thank Patryk Jedlikowski and Mikołaj Sienkiewicz for prototyping the early variants of \mname. 
  This research was supported by TAILOR, a project funded by EU Horizon 2020 research and innovation program under GA No. 952215, by the statutory funds of Poznan University of Technology and the Polish Ministry of Education and Science, grant no. 0311/SBAD/0726.
\end{acks}

\bibliographystyle{ACM-Reference-Format}
\bibliography{bibliography} 

\end{document}